\documentclass{article}

\usepackage[square,sort,comma,numbers]{natbib}

\usepackage[preprint]{neurips_2019}

\usepackage[utf8]{inputenc} % allow utf-8 input
\usepackage[T1]{fontenc}    % use 8-bit T1 fonts
\usepackage{hyperref}       % hyperlinks
\usepackage{url}            % simple URL typesetting
\usepackage{booktabs}       % professional-quality tables
\usepackage{amsfonts}       % blackboard math symbols
\usepackage{nicefrac}       % compact symbols for 1/2, etc.
\usepackage{microtype}      % microtypography
\usepackage{graphicx}
\usepackage{graphicx}
\graphicspath{ {.} }
\usepackage[toc,page]{appendix}

\title{Extreme Multi-Domain, Multi-Task Learning With Unified Text-to-Text Transfer Transformers}

\author{
  Adebayo Oshingbesan \\
  Carnegie Mellon University \\
  Kigali, Rwanda\\
  \texttt{oaadebay@andrew.cmu.edu} \\
  \AND
  Courage Ekoh \\
  Carnegie Mellon University \\
  Kigali, Rwanda\\
  \texttt{coekoh@andrew.cmu.edu} \\
  \And
  Germann Atakpa \\
  Carnegie Mellon University \\
  Kigali, Rwanda \\
  \texttt{gatakpa@andrew.cmu.edu} \\
  \And
  Yonah Byarugaba \\
  Carnegie Mellon University \\
  Kigali, Rwanda\\
  \texttt{ybyaruga@andrew.cmu.edu}
}

\begin{document}

\maketitle

\begin{abstract}
    Text-to-text transformers have shown remarkable success in the task of multi-task transfer learning, especially in natural language processing (NLP). However, while there have been several attempts to train transformers on different domains, there is usually a clear relationship between these domains, e.g.,, code summarization, where the natural language summary describes the code. There have been very few attempts to study how multi-task transfer learning works on tasks in significantly different domains. In this project, we investigated the behavior of multi-domain, multi-task learning using multi-domain text-to-text transfer transformers (MD-T5) on four tasks across two domains - Python Code and Chess. We carried out extensive experiments using three popular training strategies: Bert-style joint pretraining + successive finetuning, GPT-style joint pretraining + successive finetuning, and GPT-style joint pretraining + joint finetuning. Also, we evaluate the model on four metrics - Play Score, Eval Score, BLEU Score, and Multi-Domain Learning Score (MDLS). These metrics measure performance across the various tasks and multi-domain learning. We show that while negative knowledge transfer and catastrophic forgetting are still considerable challenges for all the models, the GPT-style joint pretraining + joint finetuning strategy showed the most promise in multi-domain, multi-task learning as it performs well across all four tasks while still keeping its multi-domain knowledge.

\end{abstract}

\footnotetext{\href{https://github.com/Dehbaiyor/IDLFall2021Project}{Github Repo}}
%%%%%%%%%%% Introduction %%%%%%%%%%%%%%%
\section{Introduction}
Teaching a machine to carry out more human-like tasks like creative thinking is a concept that goes as far as back as the 1930s, with significant progress made over the past eighty years due to big data, increased computational power, and better architectures \cite{ross_1933_machines, rosenblatt_1958_the, wikipediacontributors_2019_ai}. Chess-playing is one of the tasks that was considered a crucial measure of progress in AI. A world-champion chess-playing computer was listed as one of the AI Grand Challenges in 1995 \cite{reddy_1995_grand}. This challenge has since been achieved with computers consistently beating the best humans in chess, even in handicap matches where the chess engines start with fewer pieces \cite{wikipediacontributors_2019_deep,wikipediacontributors_2019_humancomputer, silver_2018_rl}. The same push that we saw for chess-playing computers in the 1990s is now being seen for code-writing AI \cite{dutta_2019_codenet}.

With significant progress made in teaching neural networks to learn how to perform a single task, there has been a new push in the past few years to teach one neural network model how to perform multi-task learning \cite{zhang_2017_multitask, crawshaw_2020_multitask, sanh_2021_multitask}. One architecture that has shown some promise in this area is the transformer architecture \cite{bahdanau_2014_neural, vaswani_2017_attention}. This architecture has found applications in several domains of deep learning and has been shown to be capable of zero-shot or few-shot learning on several natural language tasks \cite{zhang_2017_multitask, sanh_2021_multitask, raffel_2019_t5,  aribandi_2021_ext5}. 

In this project, we aim to assess how the multi-task learning paradigm with unified text-to-text transformers extend beyond natural language tasks into extremely different domains - chess and code. Our key research question is - will unified text-to-text transformers perform as well across tasks from multiple domains  as they have performed across NLP tasks? The following sections describe what related works have been carried out, our methods, results \& discussion, and then the conclusion.

%%%%%%%%%%%%%%%%%%%%%%%%%%%%%%%%%%%% Related Works %%%%%%%%%%%%%%%%%%%%%%
\section{Related Works}
\subsection{Deep Learning for Chess}
Several works have attempted the use of deep learning for chess-playing. DeepChess leveraged the combination of Deep Belief and Siamese Networks to build a chess engine that had a playing style resembling that of human grandmasters \cite{david_2016_deepchess}. Other researchers have tried to incorporate explainability into chess-playing machines through automated commentary to make the engines easier to understand \cite{jhamtani_2018_learning, zang_2019_automated}.

In 2019, a novel end-to-end deep learning model for chess was proposed. It leveraged the use of a sentiment-based evaluation function obtained by training on chess commentaries using an LSTM model \cite{kamlish_2019_sentimate}. The first chess transformer model was built in 2020 by finetuning the GPT-2 architecture to generate chess moves \cite{noever_2020_the}. Rather than predict moves, \cite{toshniwal_2021_learning} evaluated the ability of language models to track chess states and showed some success.

\subsection{Deep Learning for Code}
Just like chess, there have been several attempts to carry out code-like tasks such as code summarization, code generation, code-to-code translation, among others. However, unlike chess, most of the works in chess have been using the transformer framework. Several variants of the transformer model for code-related tasks such as CodeBERT, CodeGPT, CodeT5, and so on have been proposed with varying levels of success \cite{dutta_2019_codenet, toshniwal_2021_learning, ahmad2021unified, mastropaolo2021studying, wang_2021_codet5, puri_2021_codenet}. Of all these variants, only CodeT5 follows the unified text-to-text framework that we adopt in this project.

\subsection{Multi-task Learning}
Multi-task learning at scale using a text-to-text framework was popularized by the T5 architecture \cite{raffel_2019_t5}. T5 is a transformer-based architecture that uses a text-to-text approach to model varied NLP tasks such as translation, question answering, and classification as feeding the model text as input and training it to generate some target text. \cite{sanh_2021_multitask} showed that the unified text-to-text framework enabled zero-shot task generalization to multiple NLP tasks. 

ExT5 \cite{aribandi_2021_ext5} scaled up the idea of multi-task learning for NLP to a lot of tasks (over 100) and focused on multi-task pretraining rather than finetuning. Other works, such as \cite{vu2020exploring}, tried to understand how the relationship between tasks affects downstream learning in large language models between NLP tasks, highlighting the problems of catastrophic forgetting and negative transfer. 

Multi-domain learning has been previously attempted in several works \cite{lu2021codexglue, wang_2021_codet5, jhamtani_2018_learning, zang_2019_automated}. However, the domains are related, e.g., English chess commentary or code docstring generation. As far as we know, no work has considered how multi-task learning works across multiple domains that are significantly different where there is no relationship between the tasks' domains.

%%%%%%%%%%%%%%%%%%%%%%%%%%%%%%%%%%%% Baseline Models %%%%%%%%%%%%%%%%%%%
\section{Methods}
\subsection{Introduction}
The unified text-to-text framework \cite{raffel_2019_t5} provides a relatively simple way to train a single model on a wide variety of tasks using the same loss function and decoding procedure. Despite not having the advantage of specialization of task-specific architectures, this framework obtains comparable performance to task-specific architectures. 

In this research, we train several transformers models using the unified text-to-text framework with a multi-domain, multi-task objective. In particular, we pretrain the model on both code and chess data before then finetuning the model to carry out the following tasks:
\begin{itemize}
    \item Chess move generation.
    \item Chessboard state evaluation.
    \item Code generation from an English prompt.
    \item Code summarization using English language.
\end{itemize}

\begin{figure}[ht]
\includegraphics[scale = .7]{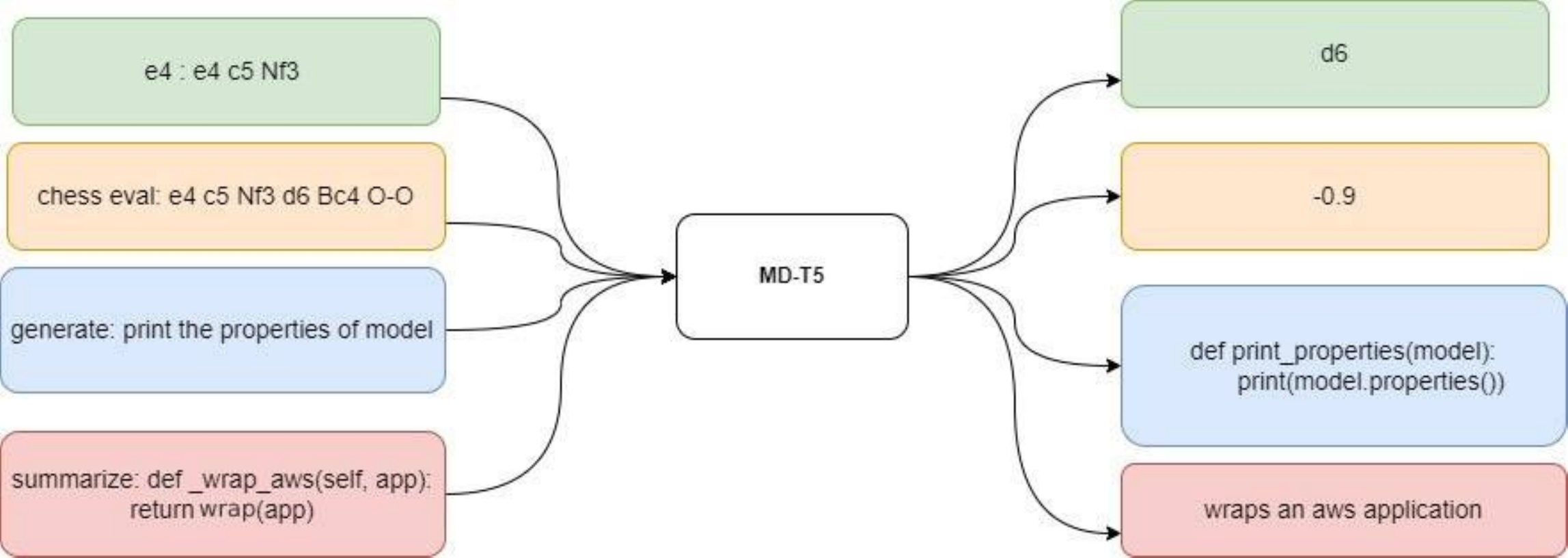}
\label{md-t5}
\caption{Multi-Domain Text-to-Text Transfer Transformer (MD-T5) Tasks}
\end{figure}

Figure \ref{md-t5} provides a pictorial summary of each of the tasks as a text-to-text problem.

The two domains were chosen based on these criteria. First, the transformer model has been applied individually to these two domains with a reasonable level of success. Second, the two domains do not have any direct or indirect relationship with one another except for the fact that they can both be modelled as text-to-text problems. Finally, the two domains have some creative element to them.

\subsection{Datasets}
\footnotetext{\href{https://www.kaggle.com/courageekoh/idl-project-code-encoder-set}{Code Dataset}, \href{https://www.kaggle.com/gatakpa/chess-evaluation-dataset}{Chess Dataset}}
We curated chess PGN data from several open-source channels such as Lichess \cite{lichessorg} and Kaggle \cite{a35, a1, chess, a60000}. At the end of our extensive dataset collection process, we ended up with 14.3 million chess PGN games and 12.7 million evaluated chess positions. This combined chess dataset is about 7x the dataset size typically used in the literature \cite{noever_2020_the, kamlish_2019_sentimate, toshniwal_2021_learning}. 

We use about 10.5 million games of the 14.3 million chess games for pretraining and the rest for finetuning for the move prediction task. Furthermore, the 3.8 million games for finetuning were split into train and test set in a 99/1 ratio. For the board evaluation task, we also used a 99/1 ratio train-test ratio during finetuning for the 12.7 million evaluated chess positions.

For the coding dataset, we rely on three well-known code datasets – CodeNet \cite{dutta_2019_codenet}, CodeSearchNet \cite{puri_2021_codenet}, and CodeXGLUE \cite{lu2021codexglue}. We extracted 1 million Python functions from the CodeSearchNet and CodeXGLUE datasets and used this during pretraining. Similarly, we extract about 350k Python code and related docstrings from the CodeNet dataset \cite{puri_2021_codenet} for finetuning and split it into train and test set in a 90/10 ratio.
    
\subsection{Models Description}
\subsubsection{AutoRegressive Language Models (GPT Family)}
Autoregressive Language Models are pretrained on the classic language modeling task in which they guess the next token having read all the previous ones. 
%They are based on RNN (e.g., Vanilla, LSTM, GRU) or based on the transformer architecture. In our work, we are interested in the latter as they achieve SOTA performance in most of the Natural Language Understanding tasks such as textual entailment, question answering, semantic similarity assessment, and document classification. 
They correspond to the decoder of the original transformer model \cite{NIPS2017_3f5ee243}, and a mask is used on top of the full sentence so that the attention heads can only see what was before in the text and not what’s after. Although these models can be fine-tuned and achieve great results on many tasks, the most natural application is text generation. A typical example of such models is GPT \cite{radford2019language}.

%% Reference original transfomer architeture and GPT architectuer papres
The GPT architecture comes from a Generative Pretraining of a language model on a diverse corpus of unlabeled text, followed by discriminative finetuning on each specific task \cite{radford2019language}. It leads to large performance gains on Natural Language Understanding tasks such as textual entailment, question answering, semantic similarity assessment, and document classification. The training procedure consists of two stages. The first is learning a high-capacity language model on a large corpus of text, and the second is the finetuning stage, where the model is adapted to a discriminative task with labeled data.

\textbf{Unsupervised Pretraining}\\
Let $U=\{u_1, \dots, u_n\}$ be an unsupervised corpus of tokens. A multi-layer transformer decoder model, which is a variant of the original transformer model, is used to maximize the following likelihood:
\begin{equation}
    \label{eqn:gtpobjective}
    L_1(U) = \Sigma_i log P(u_i|u_{i-k}, \dots, u_{i-1}; \Theta)
\end{equation}
where $k$ is the size of the context window, and the conditional probability $P$ is modeled using a neural network with parameters $\Theta$.

\textbf{Supervised finetuning} \\
After training the model with the objective in Equation: \ref{eqn:gtpobjective}, the parameters are adapted to the supervised target task with a labeled dataset $C$ with a sequence of input tokens for each instance, $x^1, \dots, x^m$, and a label $y$. The inputs are passed through the pre-trained model to obtain
the final transformer block’s activation $h^m_l$ , which is then fed into an added linear output layer with parameters $Wy$ to predict $y$.
\begin{equation}
    P(y|x^1, \dots, x^m) = softmax(h^m_lW_y)
\end{equation}
This gives the following objective to maximize:
\begin{equation}
    L_2(C) = \Sigma_{(x,y) log P(y|x^1, \dots, x^m)}
\end{equation}

\subsubsection{Masked Language Models (BERT Family)}
Autoencoding models are pretrained by corrupting the input tokens in some way and trying to reconstruct the original sentence. These models rely on the encoder part of the original transformer model allow the model to look at all the tokens in the attention heads. Masked language models randomly mask some of the tokens from the input during the pretraining, and the objective is to predict the original vocabulary id of the masked word based only on its context. The Bidirectional Encoder Representations from Transformers \cite{devlin-etal-2019-bert}, "BERT", is one example of such models. One of the key differences between the BERT model and the GPT model is that BERT uses a bidirectional Transformer while GPT uses a left-to-right Transformer. Also, the pretraining in BERT uses a Masked Language Model while GPT uses an AutoRegressive Language Model.

%%%%%

\subsection{Baselines}
Since no model has worked across the domain of chess and coding, we have several baseline models across both tasks. The first baseline is a variant of the chess transformer \cite{noever_2020_the} for chess move prediction. The second baseline is another variant of the chess transformer \cite{noever_2020_the} for board state evaluation. The third and fourth baselines are the finetuned baseline for code summarization and code generation by CodeT5 \cite{wang_2021_codet5}.

\subsubsection{The Chess Transformer}
This choice of baseline was motivated by the fact that it was the most recent published work we could find and looks to be state of the art in the field of using transformers for chess-related tasks. Rather than use the complete GPT-2 architecture as the authors did, we decided to go with a smaller GPT architecture due to memory and training time constraints. Specifically, Table \ref{tab:diff} shows where our architecture choice for the baseline differed from [17]. Also, we trained the GPT model from scratch and did not try to fine-tune the previously pre-trained weights of GPT-2 as we believed that this would yield comparable results at a shorter time. Like in GPT2, we use the Byte-Pair Encoding tokenizer.

\begin{table}[ht]
    \centering
    \begin{tabular}{ |c|c|c| } 
     \hline
     \textbf{Parameter} & \textbf{Noever et. al} & \textbf{Our Modified Baseline} \\
     \hline
     vocab size & 50,257 & 2000 \\ 
     \hline
     max length & 1024 & 500 \\
     \hline
     Number of embedding & 768 & 256 \\
     \hline
     Number of heads & 12 & 32 \\ 
     \hline
     Training steps & 30,000 & 50,000 \\
     \hline
    \end{tabular}
    \caption{Architectural Differences Between Chosen and Implemented Baseline}
    \label{tab:diff}
\end{table}

To ensure that our model only learns from very good moves sequence, we only trained the baseline model on moves from players ranked 2400 and above (rather than the entire combined dataset). The performance obtained using this modified baseline did not differ from the reported performance (9.6\% of illegal moves generated versus 10\%). 

For the chess board state evaluation, we chose to finetune the full GPT-2 architecture on about 2 million evaluated chess positions for 3,000 steps. We chose to pretrain rather than train from scratch as the first variant because we wanted to leverage the pretrained knowledge of numbers in text that the original GPT-2 model has.

\subsubsection{CodeT5}
The CodeT5 baseline choice is motivated by the fact that it is the only code language model that is currently available that uses the same unified text-to-text framework as our work. Because of how extensive our project is and the amount of compute time it will take to train this baseline, we chose to use the pretrained weights publicly provided by \cite{wang_2021_codet5} on their summarization and code generation tasks through the HuggingFace library.

\section{Experiments}
\subsection{Description}
There are two stages in this framework: pretraining and finetuning. The approach of unsupervised pretraining followed by supervised finetuning is used. Specifically, we used two architectures: the BERT-style masked language model objective and the GPT-style autoregressive language model objective. In the finetuning step, we experimented with two formulations of the text-to-text framework: finetuning the pretrained models on each task separately and  concatenating all the subtasks into one, and finetuning once. We chose these two formulations from a literature review of several multi-task learning works described in the related works section.

All our experiments were carried out using the HuggingFace library and Pytorch. We ran three sets of experiments described in the following subsections. The models generated from each of the experiment set is named MD-T5-x where x is the experiment set. We used the following parameters across all experiments:
\begin{itemize}
    \item $vocab size = 50,000$
    \item $max position embeddings = 514$
    \item $num embeddings = 768$
    \item $num attention heads = 12$
\end{itemize}

\subsubsection{Experiment Set A}
In this experiment set, we trained a Byte Level Byte-Pair Encoding (BPETokenizer) on the mixed dataset of Python functions and Chess games using a vocabulary size of 50,000. We then pretrained a Roberta masked language model, where we replace 15\% of the spans of text on a mixed dataset of both Python functions and Chess games for a total of 500,000 training steps (about 70 hours of training on NVIDIA T4 GPUs). We then proceeded to finetune this pretrained model on the four tasks as described in section 3.1 individually in succession using an encoder-decoder architecture with the encoder and decoder weights tied. Each finetuning task was trained at an average of 30 hours per task on NVIDIA T4 GPUs.

\subsubsection{Experiment Set B}
In this experiment set, we also trained a Byte Level Byte-Pair Encoding (BPETokenizer) on the mixed dataset of Python functions and Chess games using a vocabulary size of 50,000. We then pretrain a GPT-2 language model using the same vocab size, embeddings size, and attention head as the Roberta model for a total of 500,000 training steps (about 60 hours of training on NVIDIA P100 GPUs). We then proceeded to finetune this pretrained model on the four tasks described in section 3.1 individually in succession using the pretrained GPT-2 architecture. Each finetuning task was trained for an average of about 15 hours per task (120,000 steps per task) on NVIDIA P100 GPUs.

\subsubsection{Experiment Set C}
In this experiment set, we followed the same tokenizer training and model pretraining paradigm of experiment set B. However, rather than finetune on the four tasks individually, we finetuned once on a joint balanced dataset of the four tasks using the pretrained GPT-2 model from experiment set b. The joint finetuning was trained for about 15 hours (120,000 steps) on NVIDIA P100 GPUs.

\subsection{Evaluation Metrics}
We propose four different evaluation metrics that measure how well the models perform both on the different tasks and multi-domain learning. The first metric is the play score, inspired by the Elo score \cite{ferreira2013impact}, which ranks how well the chess engine plays in relation to its competitors. The second metric is the chess evaluation score which ranks how well the chess engine evaluates the board state. The third metric is the BLEU score. The BLEU score measures the quality of the code-to-text and text-to-code translation tasks. Finally, we propose a new metric which measures how well the model keeps the knowledge of the different tasks - multi-domain learning score. We describe these evaluation metrics below:

\paragraph{Play Score (PS):}
This is an aggregate rank score that incorporates several metrics (see appendix \ref{appendix:playscore}) covering how accurate the chess engine play is and how well it understands the board state in comparison to its competitors. The higher the PS, the better the model. Mathematically, the PS of a chess engine is formulated as:
\begin{equation}
    PS = 1/n\ \times \sum_{i=0}^{n} R_i
\end{equation}
where n is the number of metrics and  $R_i$ is the rank (from largest to smallest) of a chess engine among its competitors for metric $i$.

\paragraph{Evaluation Score (ES):}
This is an aggregate rank score that incorporates several metrics (see appendix \ref{appendix:eval_score}) covering how well the chess engine correctly evaluates the board states in comparison to its competitors. The higher the ES, the better the model. Mathematically, the ES of a chess engine is formulated as:
\begin{equation}
    ES = (X + Y)/2
\end{equation}
where $X$ is the rank (from largest to smallest) of a chess engine among its competitors for the regression part of the task and $Y$ is the rank (from largest to smallest) of a chess engine among its competitors for the classification part of the task.

\paragraph{Bilingual Evaluation Understudy Score (BLEU Score):}
BLEU score is a score used for comparing a candidate's translation of the text to one or more reference translations. The higher the BLEU score, the better the model. BLEU is computed using a couple of ngram modified precisions as shown below:\\
\begin{equation}
    BLEU = BP \times exp(\Sigma_{n=1}^N w_nlogp_n)
\end{equation}
where $p_n$ is the modified precision for ngram, $w_n$ is the weight between 0 and 1 for $log p_n$ and $BP$ is the brevity penalty to penalize short machine translations. 

The BP is computed as:
\begin{equation}
   BP = \{^{1 \hspace{2cm} if \hspace {1mm} c > r}_{exp(1 - \frac{r}{c})\} \hspace {0.8cm}if \hspace {1mm} c \leq r}
\end{equation}
where $c$ is the number of unigrams (length) in all the candidate sentences, and $r$ is the best match length for each candidate sentence in the corpus.

\paragraph{Multi-Domain Learning Score (MDLS)}
This is the average of the non-token mix ratio and the cross-domain recall ratio multiplied by 100 (see appendix \ref{appendix:mdl}). It is inspired by the F1 score. 

The non-token mix ratio is a precision-like component  which measures how well the model keeps the knowledge of the two domains separate after finetuning. For example, if the model was trained on the code summarization task, it should not output chess moves (eg. e4) in its summarization outputs. This is equivalent to a human answering a technical deep learning question with a Shakespearean quote just because they happen to minor in literature.  If this happens, the precision-like component of the score penalizes the model. 

Similarly, the cross-domain recall ratio is a recall-like component which measures how much of the other domain the model remembers after finetuning. Using the example of the code summarization task again, if a model is prompted with a sequence of chess moves, we expect it still outputs valid chess chess moves. This is to ensure that the model has not lost its multi-domain knowledge even if it may be better at one domain than the other. Humans do not forget all their previous knowledge of a domain when they acquire a new knowledge of another domain even if the new domain is significantly different.

We find the harmonic mean of these two components for the multi-domain learning score. The higher the score, the better the model.

\begin{equation}
    2 \times \frac{(Non-token\ mix\ ratio)\ *\ (cross-domain\ recall\ ratio)}{(Non-token\ mix\ ratio)\ +\ (cross-domain\ recall\ ratio)}\ 
\end{equation}

%%%%%%%%%%%%%%%%%%%%%%%%%%%%%%%%%%% Results %%%%%%%%%%%%%%%%%%%%
\section{Results and Discussion}
Table \ref{metrics} shows the results across each metric for all the tasks for all the experiment sets. Appendices \ref{appendix:chesstask1} to \ref{appendix:mdl_results} shows the raw sub-metrics scores that were aggregated to obtain the overall metric scores. Appendix \ref{sample_outputs} show some sample outputs from the MD-T5 models for both the chess-related and code-related tasks. On a high level, Table \ref{metrics} shows MD-T5-B models perform better than the baseline on three out of four tasks with additional multi-domain knowledge. Similarly, MD-T5-C models outperform the baseline on two out of four tasks with significantly better multi-domain knowledge score. 

\begin{table}[!ht]
    \centering
    \caption{Results of Extreme Multi-Domain, Multi-Task Learning Experiment}
    \label{metrics}
    \begin{tabular}{|c|c|c|c|c|}
    \hline
        Metric & Baseline & MD-T5-A & MD-T5-B & MD-T5-C \\ \hline
        Play Score & 3 & 1 & \textbf{3.2} & 2.8 \\ \hline
        Eval Score & 1.5 & 2.5 & \textbf{3.5} & 2.5 \\ \hline
        CS BLEU & 20.36 & 10.79 & \textbf{31.64} & 28.9 \\ \hline
        CG BLEU & \textbf{41.48} & 11.42 & 31.37 & 30.81 \\ \hline
        MDLS & 0 & 5.77 & 13.3 & \textbf{95} \\ \hline
    \end{tabular}
\end{table}

From appendix \ref{appendix:playscore}, we see a variety of behavioral differences between the different MD-T5 models and the baseline on the move prediction task. First, the MD-T5-B model tends to generate much longer games than the other MD-T5 variants and the baseline. Furthermore, it tends to keep track of the board state for much longer on average (76 moves vs 1/54/68 moves). Despite its ability to keep track of the board state much longer, it generally plays very accurate moves, outperforming the baseline and having similar performance to the MD-T5-C model that generates much shorter games (and thus have to play much easier moves). However, all MD-T5 models struggle with ending the game and this could be attributed to the fact that they lose track of the game state as they get to the end of the game. The MD-T5-A model struggled with the task immensely.

From table \ref{metrics}, we see that all MD-T5 models outperform the baselines on the board state evaluation task. However, we note that all the models, including the baseline, perform poorly on this task. Appendix \ref{appendix:chesstask2} shows that while MD-T5-A can generate numerical values correctly when a numerical value is required 70\% of the time, its mse and accuracy values show that it still struggles with the task. The story is quite similar for MD-T5-B and MD-T5-C models, even though MD-T5-B makes better predictions for both numerical and non-numerical values compared to the other ones. Since this is a joint regression and classification problem modeled as a text-to-text problem, these results are not entirely surprising.

Table \ref{metrics} shows the MD-T5-B and MD-T5-C models significantly outperform the baseline. We posit that this is because these models were trained on just Python code while the baseline model was trained on multiple programming languages and negative transfer \cite{vu2020exploring} may have occurred. Yet, this result is impressive as shown by three of the best performing outputs provided (appendix \ref{sample_outputs}). One could even argue that the summarization provided by the MD-T5-B and MD-T5-C models for input 1 and input 2 is better or just as good as the target given the function. This is much more impressive given the fact that these models were never pretrained or even finetuned on English language data. Yet, they were able to generate fluent, concise, and relevant summaries.

While the MD-T5-B and MD-T5-C models could not outperform the baseline on the code generation task, they generate reasonable function names and code structure given the text prompt (appendix \ref{sample_outputs}). This is pretty impressive given that even a human programmer would show the similar behavior given no other context except a text prompt. Again, we posit that while the training across multiple languages could have been a disadvantage for the code summarization task, it was probably an advantage for the baseline model here as knowledge transfer would be helpful on the challenging task of code generation. Furthermore, the baseline model was trained for much longer (96 hours on a cluster of A100 GPUs) than our MD-T5 models and we know that longer training time is an additional advantage for model performance on complex tasks.

Perhaps the vital aspect of this analysis is multi-domain knowledge as measured by the multi-domain learning score. Once again, the MD-T5-A model does not yield outstanding performance while MD-T5-B and MD-T5-C models post pretty impressive results with little or no token mix (Appendix \ref{token-mix}). However, MD-T5-B models are susceptible to catastrophic forgetfulness  \cite{vu2020exploring} as seen from the cross-domain recall scores (Appendix \ref{recall-result-appdx}). This is because the finetuning is done separately rather than together as in MD-T5-C. Appendix \ref{prompting} shows typical recall results from MD-T5-B after prompting. From the prompting outputs provided, we can see that MD-T5-B suffers from catastrophic forgetting even though it still maintains some multi-domain knowledge if the prompting is done sufficiently long enough. This makes the results of MD-T5-C much more impressive as it was able to perform almost as well as MD-T5-B while still maintaining a much substantial part of its multi-domain knowledge.

The MD-T5-A model struggled with generating any meaningful performance after finetuning despite having a reasonably good performance at the pretraining stage. We believe this poor performance could be a result of two things - negative knowledge transfer and inadequate training time. Negative knowledge transfer is a known phenomenon in transfer learning typical of multi-task learning when the tasks are dissimilar \cite{vu2020exploring, chen_2021_evaluating}. While we see some negative knowledge transfer in the GPT-style models (MD-T5-B and MD-T5-C), we believe the denoising objective is why the effects are much more profound in the BERT-style model. This brings us to the next point - inadequate training time. While we trained all the models for approximately the same amount of time since they had roughly the same number of parameters, it is possible that the BERT-style model requires more training time to generate good results.

A comparison of the GPT-style models shows that they hold up well against some solid baselines. The MD-T5-B model seems to outperforms other MD-T5 models on each task across the domain. However, as we saw from the prompting examples and the cross-domain recall score, this came at the cost of forgetting most of what it had learned from the other domain i.e. losing its multi-domain knowledge. Given this context, it shows how impressive the MD-T5-C model performance is. Not only is it able to perform each task reasonably well, but it also does this while still keeping it multi-domain knowledge. Thus, the GPT-style joint pretraining and joint finetuning framework is the most promising direction for multi-domain, multi-task learning using a unified text-to-text transfer transformer architecture.

\section{Conclusion}
We investigated the  abilities of transformers to perform well across tasks from multiple domains using a unified text-to-text framework. We curated datasets from several sources and then carried out several experiments using three different training strategies: Bert-style joint pretraining + successive finetuning, GPT-style joint pretraining + successive finetuning, and GPT-style joint pretraining + joint finetuning. We chose these strategies as they were the most common in related works.

Our experiments and analysis show that the multi-domain text-to-text transfer transformer framework that we propose compares well on the individual tasks across multiple domains against powerful transformer baseline models. Furthermore, we see that the joint pretraining and finetuning framework in experiment set C performs well on individual tasks while still keeping its multi-domain knowledge. While these results are encouraging, it is still limited by the fact that it seems that we lose multi-domain knowledge when we do not finetune the tasks jointly, a strategy that is not as popularly adopted in real-world scenarios as the first two.

This research has focused on just three training strategies. However, there are a lot more strategies, even if they are far less popular \cite{raffel_2019_t5, aribandi_2021_ext5}. Thus, one natural extension to this project is the application of one or more of these other strategies to the task of extreme multi-domain learning. Similarly, this project was limited by available compute resources and it would be interesting to see how performance changes with more compute resources and training time. Finally, it will also be interesting to scale up this work to tens of extremely different domains and hundreds of tasks and see what happens then or at a smaller scale, increasing the number of tasks significantly across two or three domains.

\bibliographystyle{unsrt}
\bibliography{main}

\newpage

\begin{appendices}
\section{Play Score Sub-Metrics Description}
\label{appendix:playscore}
\paragraph{The Proportion of Illegal Chess Moves Generated:}
An illegal move in chess is any move that violates the standard rules of chess. Using the python-chess library [32], we can check to see if any generated chess move by the model is illegal. The proportion of illegal moves generated is therefore the percentage of illegal moves of all the moves generated during gameplay.

\paragraph{Average Move Number of Illegal Move Generation:}
In general, we expect the model to perform well in the opening and middle game phases (move 1 - 50) since it is likely to have seen similar positions several times in the training dataset. However, we expect it to struggle in the endgame where many positions are unique and even top human chess players struggle in the phase. Thus, the average move number of illegal move generation is to average move number where the model makes an illegal move, and it will help to keep track of how well the model plays in the end game which is a measure of how well the model understands the game of chess.

\paragraph{The Proportion of Missed End State:} 
An end state in chess is either a white win (1-0), a black win (0-1), or a draw (½ – ½). Using the python-chess library [32], we can keep track of the game state and ascertain when one of the following end states has occurred. We expect that the model can keep track of these too and generate one of the end states tokens if the game is over. The proportion of missed end state is therefore the percentage of games in which the model keeps generating chess moves even after the game should have ended.

\paragraph{Average Centipawn Loss:}
Centipawn loss is a numerical score given by a chess engine (usually StockFish) to the difference between the move you played against the strongest move available at that time. Since conventional chess engines are way better than even humans at chess, it is used as a benchmark of how well a chess player plays. A strong model should have an average centipawn score close to 0.

\paragraph{Game Length:} In general, better chess playing involves being able to play longer games as this implies that you are not losing quickly. This is particularly important for chess engines. The game length is the number of moves that the chess move prediction model can generate on average.

\section{Eval Score Sub-Metrics Description}
\label{appendix:eval_score}
\paragraph{The Ratio of Correct Numerical Values:}
The chess move evaluation is an evaluation function used to heuristically determine the relative value of a position. It's usually a real number that we decided to bound between -10 and +10 and bin in 44 bins. However, it can also be a string from a finite set of strings, when it's possible to force a mate or a draw in a few numbers of moves. Thus, we compute the fraction of time the model predicts a numerical  token (cast as a string, considering the text-to-text framework) when the true evaluation is also a numerical token.

\paragraph{The Ratio of Correct Non-numerical Values:}
This metric is conceptually the same as for the "Ratio of correct numerical values", except that here we compute the fraction of time the model predicts a non-numerical token (cast as a string, considering the text-to-text framework) when the true evaluation is also a non-numerical token.

\paragraph{Mean Squared Error:}
The chess board state evaluation is usually a real number that we decided to bound between -10 and +10 and bin in 44 bins. The motivation was to cast the regression part of the problem into a simpler classification problem given the already complex nature of the data. Then we use the mean squared error to compute the divergence between the predictions of the models and the true evaluation of the board state.

\paragraph{Accuracy:}
The accuracy is one metric for evaluating classification models, and it's the fraction of correct non-numerical chess board state evaluation that the model gets correctly. The chess move evaluation is sometimes a string, from a finite set of strings, instead of a numerical value. So we computed the accuracy to find the fraction of time the model gets the non-numerical values correctly.

\section{Multi-Domain Learning Sub-Metrics Description}
\label{appendix:mdl}
\paragraph{Non-Token Mix Ratio(NMR):}
This is the ratio of time the model generates a chess token in a code-related task and vice-versa of all the text generated. The higher the non-token mix ratio, the better the model. Mathematically, the non-token mix ratio for a model is given as
\begin{equation}
    NMR = 1 - \frac{|A\cap\ B|\ }{|A\ \cup\ B|}
\end{equation}
where A is the tokens from chess-related tasks, and B is the tokens from code-related tasks generated by that model.

\paragraph{Cross-Domain Recall Ratio (CRR):}
This is the ratio of time the model successfully returned tokens from another domain when prompted with tokens from that domain after being finetuned on a separate domain.

\section{Raw Sub-Metrics Scores for Chess Move Prediction}
\label{appendix:chesstask1}
Table \ref{movepredresults} shows the results from the chess move prediction task. The numbers in the bracket represent the metric number given the fact that each game was limited to just 70 games as this is the 90th percentile of the game length of the training dataset.

\begin{table}[ht]
    \centering
    \caption{Metrics for Chess Move Prediction Task}
    \label{movepredresults}
    \scalebox{0.75}{
    \begin{tabular}{ |c|c|c|c|c|c| } 
     \hline
     \textbf{Metric} & \textbf{Baseline}& \textbf{Experiment Set A} & \textbf{Experiment Set B} & \textbf{Experiment Set C} \\
     \hline
     \textbf{Proportion Of Illegal Moves Generated}  & 9.50\% (9.48\%) & 100\% (100\%)& 40.1 (16.5\%) & 60\% (29.7\%)\\ 
     \hline
     \textbf{Average Move Number of Illegal Move Generation} & 54 (54) & 1 (1) & 76 (52) & 68 (41) \\
     \hline
     \textbf{Average Centipawn Loss} & 1.6 (1.11) & - & 1.2 (0.74) & 0.64 (0.63) \\
     \hline
     \textbf{The Proportion of Missed End State} & 0.01\% & - & 4.6\% (2.4\%) & 0.84\% (0.28\%)\\ 
     \hline
     \textbf{Average Number of Moves in Game} & 69 & - & 101 & 55 \\
     \hline
    \end{tabular}}
\end{table}

\section{Raw Sub-Metrics Scores for Chess Board State Evaluation}
\label{appendix:chesstask2}
Table \ref{evalpredresults} shows the results from the chess move evaluation task.
\begin{table}[ht]
    \centering
    \caption{Metrics for Chess Evaluation Prediction Task}
    \label{evalpredresults}
    \scalebox{0.8}{
    \begin{tabular}{|c|c|c|c|c|}
    \hline
        \textbf{Metric} & \textbf{Baseline}& \textbf{Experiment Set A} & \textbf{Experiment Set B} & \textbf{Experiment Set C} \\ 
        \hline
        \textbf{Ratio of correct numerical values}  & 0.2\% & 70\% & 50.6\% & 51.8\% \\ 
        \hline
        \textbf{Ratio of correct non-numerical values} & 10.3\% & 2.2\% & 7.3\% & 7.1\% \\ 
        \hline
        \textbf{Mean Squared Error} & 2.42 & 60.59 & 43.42 & 76.45 \\ 
        \hline
        \textbf{Accuracy} & 0\% & 0.0\% & 26.02\% & 25.35\% \\ 
        \hline
    \end{tabular}`
    }
\end{table}

\section{Raw Sub-Metrics Scores for Multi-Domain Learning}
\label{appendix:mdl_results}
Table \ref{token-mix} shows how many times across each experiment set did the model not introduce a token from another domain to a specific domain. 
\begin{table}[ht]
    \centering
    \caption{Average Non-Token Mix Ratio Across Experiment Sets}
    \label{token-mix}
     \scalebox{1.0}{
    \begin{tabular}{|c|c|c|c|}
    \hline
        ~ & \textbf{Experiment A} & \textbf{Experiment B} & \textbf{Experiment C} \\ 
        \hline
        \textbf{Non-Token Mix Ratio} & 96\% & 0\% & 0.001\% \\ 
        \hline
    \end{tabular}
    }
\end{table}

Table \ref{recall-result-appdx} shows how many times across each experiment set did the model successfully recall cross-domain knowledge. 
\begin{table}[!hb]
    \centering
    \caption{Average Cross-Domain Recall Ratio Across Experiment Sets}
    \label{recall-result-appdx}
     \scalebox{1.0}{
    \begin{tabular}{|c|c|c|c|}
    \hline
        ~ & \textbf{Experiment A} & \textbf{Experiment B} & \textbf{Experiment C} \\ 
        \hline
        \textbf{Cross Domain Recall Ratio} & 10.4\% & 7.1\% & 90.6\% \\ 
        \hline
    \end{tabular}
    }
\end{table}

\section{Sample Outputs from MD-T5 Models}
\label{sample_outputs}
\begin{figure}[ht]
\centering
\includegraphics[scale = 0.45]{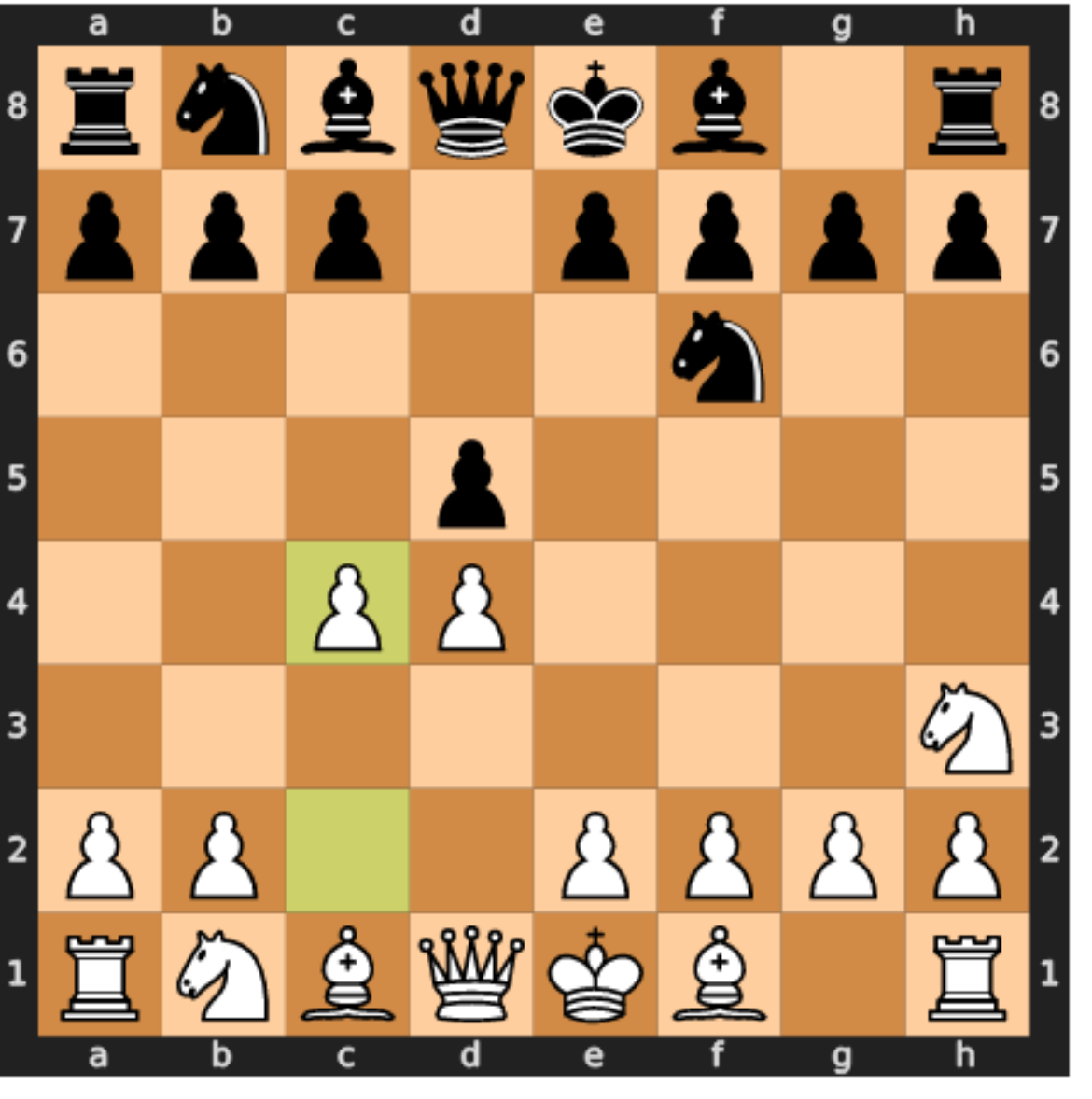}
\includegraphics[scale = 0.45]{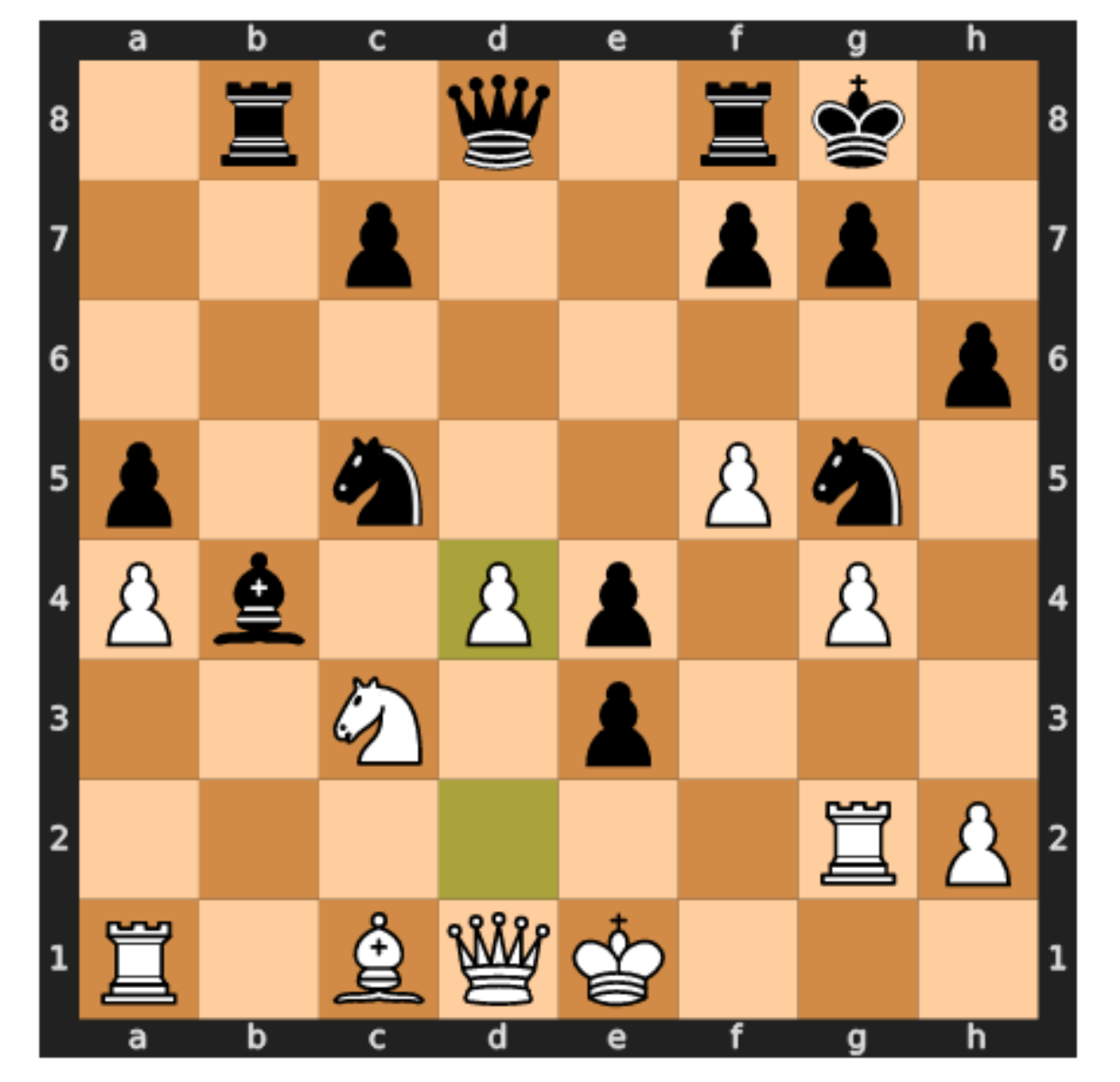}.
\includegraphics[scale = 0.45]{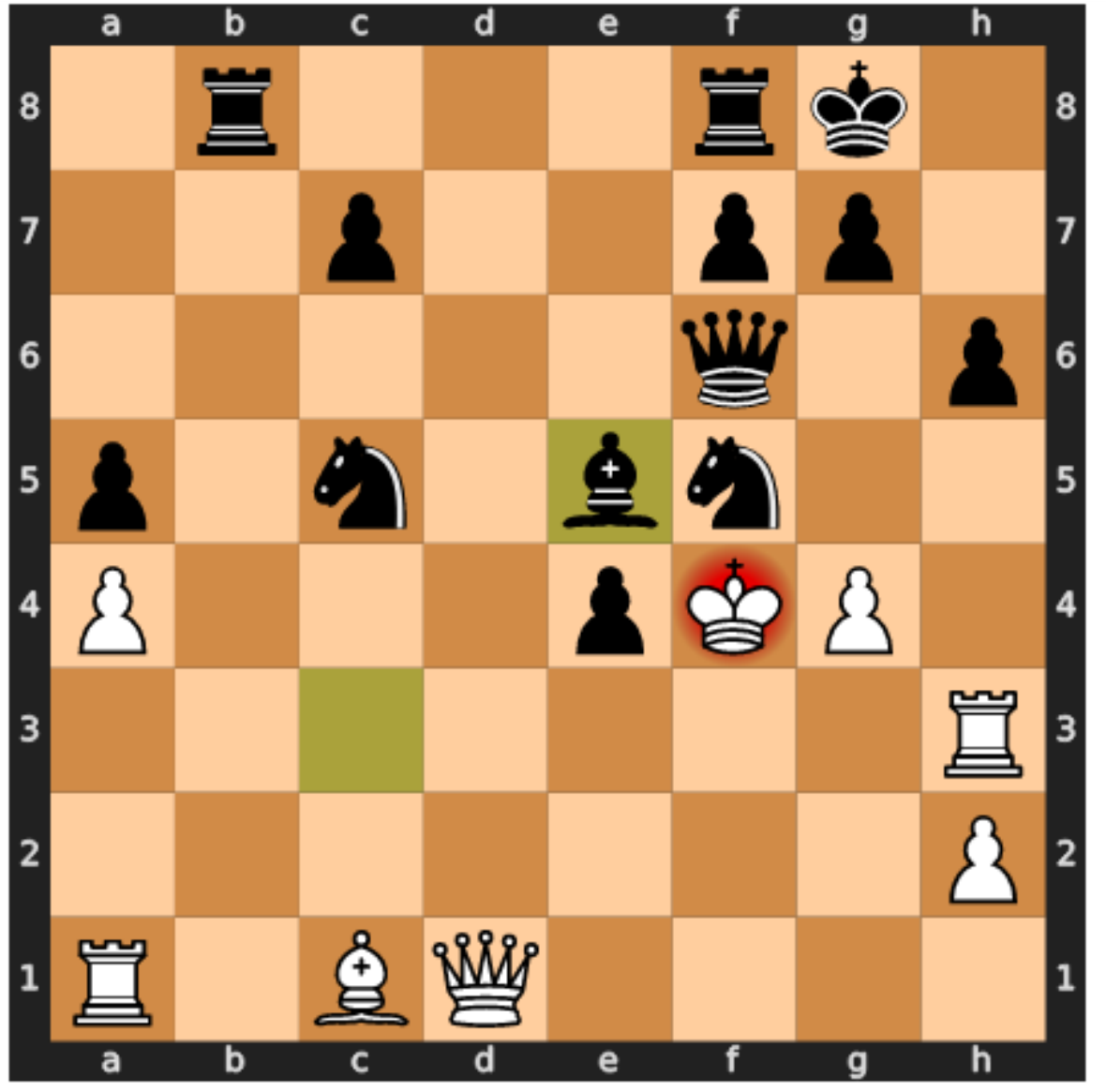}
\label{chess-open}
\caption{Sample Positions of Game Between A MD-T5 Model And Stockfish 14}
\end{figure}

Figure \ref{chess-open} provides some sample positions from a chess game played between the strongest MD-T5 variant and Stockfish-14, the best chess engine currently available.

Here are three of the best performing outputs from the code summarization tasks:
\begin{itemize}
    \item 
        \textbf{input 1}: \begin{verbatim} summarize: 
        def get_courses_for_regid(self, regid, params={}):
        """
        
        """
            self._as_user = regid
            data = self._get_resource("/api/v1/courses", params=params)
            self._as_user = None
    
            courses = []
            for datum in data:
                if "sis_course_id" in datum:
                    courses.append(CanvasCourse(data=datum))
                else:
                    courses.append(self.get_course(datum["id"], params))

        return courses
         \end{verbatim}
         
         \textbf{target:} Return a list of courses for the passed regid. \newline
         
        \textbf{experiment A:} 
        \begin{verbatim}
            s and for to the a an of from 
            in_.].'].').'),'))'])()) )))))])]))]),],'],']']:]::]]
            ()((_(( (destpathname1'2, default
        \end{verbatim}
         
         \textbf{experiment B:} Returns the list of courses for the given regid \newline
         
         \textbf{experiment C:} Return a list of courses for the given regid. \newline
         
             \item 
        \textbf{input 2}: \begin{verbatim} summarize: 
        def save_authorization_code(self, client_id, code, request,
                                *args, **kwargs):
        """"""
            log.debug(
                'Persist authorization code %r for client %r',
                code, client_id
            )
            request.client = request.client or self._clientgetter(client_id)
            self._grantsetter(client_id, code, request, *args, **kwargs)
            return request.client.default_redirect_uri
         \end{verbatim}
         
         \textbf{target:} Persist the authorization code.. \newline
         
        \textbf{experiment A:}
        \begin{verbatim}
            s the a an of to for and or from in_.2pathnameinfodebugwarningwarnerrorexceptionresult
            resultsfileslineslineaxis axis column 
            featurefeatureedgeedgesGraphgraph graph G A K D Q H E M S
        \end{verbatim}
         
         \textbf{experiment B:} Handle authorization code. \newline
         
         \textbf{experiment C:} Save the authorization code. \newline
   
           \item 
        \textbf{input 3}: \begin{verbatim} summarize: 
        def instruction_list(self):
            """
            """
            instruction_list = []
            for instruction in self.data:
                if isinstance(instruction, CompositeGate):
                    instruction_list.extend(instruction.instruction_list())
                else:
                    instruction_list.append(instruction)
        return instruction_list
     \end{verbatim}
     
     \textbf{target:} Return a list of instructions for this CompositeGate. \newline
     
    \textbf{experiment A:}
    \begin{verbatim}
        =. thename _ _ to allsr this an of((' ( on a not, andThe0 in or is 
        from for Retrieve None self get str model methodzes
        CalculateRunssummarize through format c -
    \end{verbatim}
     
     \textbf{experiment B:} Returns a list of instructions for the given instruction. \newline
     
     \textbf{experiment C:} Returns a list of ``ListingCompositeGate`` instructions. \newline
   
\end{itemize}

Here are three of the best performing outputs from the code summarization tasks: \newline

    \textbf{input 1:} generate: Returns True if $node\_type ==$ value. \newline
        
    \textbf{target:} \begin{verbatim}
     def is_type(self, value):
        """
        """
        if isinstance(value, tuple):
          for opt in value:
            if self.node_type == opt:
              return True
          return False
        else:
        return self.node_type == value
    \end{verbatim} 
 
    \textbf{experiment A:}
    \begin{verbatim}
    "):()):())])]))])))])).)].'].'][''][__[[_=[=[')['))'))
    clsself  b a Generate Update Loadcul Print Cal Find Mark
    RunheadlineUSRURATIONtestgraphsrcpathlector pexpect Append 
    \end{verbatim}
    
    \textbf{experiment B:} 
    \begin{verbatim}
    def get_value(self, node_type, value): 
        """ 
        """ 
        if node_type == value: 
            return False 
        if node_type == value: 
            return True 
        if node_type == value: 
            return True 
        return False
    \end{verbatim}
    
    \textbf{experiment C:}
    \begin{verbatim}
    def input_type(self, val): 
        """ 
        
        """ 
        if self._node_type == "node_type": 
            return True 
        if val.startswith("node_type"): 
            return False 
        if val.startswith("node_type"): 
                return True 
        return False
    \end{verbatim}

    \textbf{input 2:} generate: Returns names of features. If features is None, returns all    features. Otherwise assumes the user is trying to find the order of the    features. \newline
        
    \textbf{target:} \begin{verbatim}
    def get_feature_names(self, features=None):
        """   """
        if features:
            return self.feature_table.get_ordered_names(features)
        else:
            return self.feature_table.feature_names
    \end{verbatim}
 
    \textbf{experiment A:}
    \begin{verbatim}
    processdef LOG Log                                                        
    b d a h g e exc err error elif if because
    somethinggwverificationwiflaceongs Find Generate Par directly give`.``.
    Perform
    generate """""" """""")'))):
    \end{verbatim}
    
    \textbf{experiment B:} 
    \begin{verbatim}
    If features is None, then features are returned. 
    def get_features(self, features): 
        """ 
        """ 
        features = self.get_features(features) 
        return [features[0] for features in self.features]
    \end{verbatim}
    
    \textbf{experiment C:}
    \begin{verbatim}
    This features is created to the features. 
    def features(self): 
        """ 
        """ 
        if self.current_features is None: 
            return self.current_features 
        else: 
            return self.current_features
    \end{verbatim}

    \textbf{input 3:} generate: Gets the model data associated with the given name.        If it was loaded, a cached copy will be returned.    It it was not loaded, it will be loaded and cached. \newline
        
    \textbf{target:} \begin{verbatim}
    def getModelData(self,name):
        """
        
        """
        if name in self.modelcache:
            return self.modelcache[name]
        return self.loadModelData(name)
    \end{verbatim}
 
    \textbf{experiment A:}
    \begin{verbatim}
    "):_):_) f d a multiple another Find Generate Return Loadcul
    generatecellcountidindexis2node node._. list string ''' """""" """""")'))                                                                           
    
                       traceaux std server
    \end{verbatim}
    
    \textbf{experiment B:} 
    \begin{verbatim}
    def get_model_data(self, name, cached_name, cached_name, cache_name): 
        """ 
        """ 
        if name in self.model_data: 
            return self.model_data[name] 
        else: 
            return self.model_data[name]
    \end{verbatim}
    
    \textbf{experiment C:}
    \begin{verbatim}
    def get_model_data_loaded_name(self, name): 
    """ 
    """ 
    if not self.is_valid(): 
        return None 
    return self.get_model(name)
    \end{verbatim}

\section{Prompting for MD-T5-B Models}
\label{prompting}
Prompt 1 was given to the chess move predicted model. Prompt 2 was given to the chess board state evaluation model. Prompt 3 was given to the code summarization model. Prompt 4 was given to the code generation model. Here are samples of the typical results we obtained:

\begin{verbatim}
    prompt 1: def sharpe_ratio(return_series, N = 255, rf = 0.01, annualized = True):
    
    output 1: 2 Rf5+ Ke3 a6 a4 c6 a5 Rf7 b3 Raf8 b4 g5 g4 Kg7 ...
    
    prompt 2: def _get_abs_corr_mat(self, X_filled, tolerance): n_features = X_filled.shape[1]
    
    output 2: 7 g3 Nb6 Bg2 Nc6 O-O Be6 b3 Qd7 Bb2 Bh3 ... -1.6
    
    prompt 3: e4 e5 Nf3 Nc6 Bb5 f6 O-O fxe5 Bxc6 dxc6 Nxe5 Bd6 
    
    output 3: 
    Request. Please decimality with an E5KE KEY "
                                "the function at base64 b64 b64 KEY "
                                "for full functions.")
    if e50 <= e50 <= e50:
        raise ValueError("E5001 must be 5000.")
    if e50 <= e50 <= e50 <= e50 <= e50:
        raise ValueError("E5001 must be 500.")
    if e50 <= e50 <= e50 <= e50 <= e50 <= e50:
        raise ValueError("E5001 must be 5001 must be 500.")
    if e50 <= e50 <= e50 <= e50:
    
    prompt 4: e4 e5 Nf3 Nc6 Bb5 f6 O-O fxe5 Bxc6 dxc6 Nxe5 Bd6 d3 Ne7 c3 Ng6 Nxg6
    
    output 4: 
    O4 15
    failed_e2 = FAILE(failed_e2, buffer)
    if failed_e2.has_effective():
        prb.set_effective(failed_e2, buffer)

    # Setting all events for a series to make sure that
    all_effective_effective_effective_effective_effective_effective_effective_effectives(
        failed_event, buffer)

    # Parse that the reverse times to prevent only
    next_eff
\end{verbatim}
\end{appendices}

%%%%%%%%%%%%%%%%%%%%%%%%%%%%%%%%%%%%

\end{document}